\documentclass{article}
\usepackage{arxiv}
\usepackage[spanish,es-tabla]{babel} 
\usepackage[utf8]{inputenc} 
\usepackage[T1]{fontenc}    
\usepackage[colorlinks=true, allcolors=blue]{hyperref}
\usepackage{url}            
\usepackage{booktabs}       
\usepackage{amsmath}
\usepackage{amsfonts}       
\usepackage{nicefrac}       
\usepackage{microtype}      
\usepackage{lipsum}
\usepackage{graphicx}
\usepackage{xcolor}
\usepackage{makecell}
\usepackage{float}  
\usepackage[labelfont=bf]{caption} 

\usepackage{tikz} 
\usetikzlibrary{babel}
\usetikzlibrary{shapes.geometric, shapes.symbols, arrows.meta, shapes.multipart, arrows.meta, positioning, fit, backgrounds, calc}

\tikzstyle{startstop} = [circle, draw=black]
\tikzstyle{process} = [rectangle, minimum width=4cm, minimum height=1cm, text centered, text width=3.75cm, draw=black]
\tikzstyle{decision} = [diamond, minimum width=3cm, minimum height=1cm, text centered, draw=black]
\tikzstyle{arrow} = [thick,->,>=stealth]

\definecolor{bluebox}{RGB}{0,102,204}

\tikzset{
	box/.style={
		rectangle, draw=black, thick, fill=white, 
		minimum width=3cm, minimum height=1.5cm, align=center
	},
	arrow/.style={
		->, >=Stealth, thick, bluebox
	},
	label/.style={
		font=\scriptsize\sffamily, align=center
	}
}

\title{Geolog-IA: Sistema Conversacional sobre Tesis Académicas}

\author{
\large Micaela Fuel Pozo
\And
\large Andrea Guatumillo Saltos
\And
\large Yeseña Tipan Llumiquinga
\And
\large Kelly Lascano Aguirre
\And
\large Marilyn Castillo Jara
\And
\large Christian Mejía-Escobar
\\
\\
\texttt{\{jmfuel,adguatumillo,yytipan,kelascanoa1,mdcastilloj,cimejia\}@uce.edu.ec}
\\
\\
Facultad de Ingeniería en Geología, Minas, Petróleos y Ambiental (FIGEMPA)\\
Universidad Central del Ecuador\\
Quito, Ecuador\\
}

\begin{document}
\maketitle
\begin{abstract}

Este estudio presenta el desarrollo de Geolog-IA, un novedoso sistema conversacional basado en inteligencia artificial que responde de manera natural a preguntas sobre las tesis de Geología de la Universidad Central del Ecuador.
Nuestra propuesta emplea los modelos de lenguaje Llama 3.1 y Gemini 2.5, que se complementan con una arquitectura de Generación Aumentada por Recuperación (RAG) y una base de datos SQLite.
Esta estrategia permite superar problemas como las alucinaciones y la desactualización del conocimiento.
La evaluación del desempeño de Geolog-IA con la métrica BLEU alcanza un promedio de 0.87, lo que indica una alta coherencia y precisión en las respuestas generadas.
El sistema ofrece una interfaz intuitiva y disponible en la web, lo que facilita la interacción y recuperación de información para directivos, docentes, estudiantes y personal administrativo de la institución.
Esta herramienta puede ser un apoyo clave en la educación, formación e investigación y establece una base para futuras aplicaciones en otras disciplinas.

\end{abstract}


\section{Introducción}
\label{sec:intro}
En un mundo cada vez más interconectado y competitivo, la popular frase ``la información es poder'' adquiere mayor relevancia en cualquier contexto de nuestra vida.
Disponer de información útil y oportuna es un factor clave para alcanzar los objetivos y lograr el éxito personal, académico y profesional.
En el ámbito educativo, una de las fuentes más valiosas de información son las tesis de titulación, las cuales son producto del trabajo conjunto de docentes y estudiantes para profundizar en temas especializados, investigar y generar conocimiento.


En muchas ocasiones, la accesibilidad y la facilidad de uso de esta información se ven afectadas por sistemas de búsqueda y difusión ineficientes y poco amigables. Este tipo de sistemas sigue siendo el mecanismo principal que disponen muchas instituciones educativas \cite{ref12}.
Tomando como un caso de estudio a la Carrera de Geología de la Universidad Central del Ecuador, el acceso a las tesis de titulación se realiza por medio de bibliotecas físicas, repositorios digitales y buscadores en línea. Estos mecanismos presentan ciertas barreras que impiden el total aprovechamiento de estos importantes documentos académicos.

Por una parte, las tesis se almacenan de forma física en bibliotecas. Para su acceso, los usuarios deben obtener permisos, llenar formularios, revisar catálogos y realizar una búsqueda manual rigiéndose a los horarios de atención que no siempre coinciden con su disponibilidad. Este proceso resulta laborioso y demanda una considerable inversión de tiempo.
Por otra parte, el repositorio digital basado en la plataforma \textit{dspace} \cite{ref31} permite gestionar y difundir las tesis; sin embargo, la interfaz es poco intuitiva, lo que dificulta su uso y complica el proceso de búsqueda. Además, el sistema se encuentra fuera de servicio en muchas ocasiones.
También, los motores de búsqueda como \textit{Google} permiten acceder a las tesis, pero es necesario colocar palabras clave específicas, lo que puede devolver resultados irrelevantes, obligando al usuario a revisar manualmente múltiples fuentes para extraer la información deseada, consumiendo mucho tiempo y esfuerzo.

Por tanto, surge la necesidad de desarrollar una solución efectiva que proporcione un acceso directo y permanente a estas tesis, optimice el tiempo y esfuerzo en la búsqueda y análisis de múltiples fuentes, y que permita una interacción fácil y natural con el usuario.
El presente trabajo propone el uso de técnicas de procesamiento del lenguaje natural (NLP), un campo de la Inteligencia Artificial (IA) que 
ha experimentado un avance vertiginoso en su afán de que las computadoras puedan entender preguntas y proporcionar respuestas de la misma forma en la que los seres humanos se comunican \cite{ref25}. 
En este contexto, el aprendizaje automático profundo ha sido la base para el entrenamiento y la generación de modelos de lenguaje extensos (LLM), que están basados en arquitecturas avanzadas de redes neuronales.

A pesar de las sorprendentes capacidades de los LLMs actuales, su uso puede presentar ciertos inconvenientes como alucinaciones y desactualización del conocimiento \cite{ref32}.
Para enfrentar estas limitaciones, los documentos de tesis podrían utilizarse para realizar un \textit{fine-tuning} o ajuste del modelo; sin embargo, este proceso es muy complejo y costoso \cite{ref33}.
Por esta razón, se emplea la técnica RAG (Retrieval-Augmented Generation) \cite{ref29}, que permite combinar la capacidad de un LLM para entender y generar lenguaje natural con la recuperación de información específica, precisa y actualizada desde una fuente externa como una base de datos SQL (Structured Query Language) \cite{ref26}.

Nuestro objetivo es implementar un sistema conversacional basado en LLM y RAG-SQL que facilite el acceso, análisis y extracción de información relevante de las tesis de titulación. Esta herramienta podrá beneficiar a todos los miembros de la institución.
Los docentes pueden evaluar el desempeño estudiantil, tener una guía para la actualización de contenidos y materiales didácticos, mejorar la práctica docente y orientar a sus estudiantes en la elección de temas novedosos para futuras tesis. 
Este recurso proporciona a los estudiantes una hoja de ruta clara para sus investigaciones, evitando la tediosa revisión manual de numerosas fuentes. Además, facilita estructurar y redactar eficazmente sus proyectos de investigación y trabajos de titulación.
Los directivos pueden obtener de manera rápida y concreta información valiosa para verificar el cumplimiento de la misión, visión y objetivos de la Carrera, así como una toma de decisiones sobre el desarrollo estratégico de la institución para mejorar la calidad y excelencia de la formación académica.
También, el personal administrativo puede automatizar la búsqueda de información; los procesos de documentación se agilizarán considerablemente, permitiendo que los recursos humanos se enfoquen en tareas más estratégicas y menos repetitivas.

Consecuentemente, el sistema propuesto representa una herramienta clave para cada participante del equipo institucional. Su implementación no solo optimizará el uso del tiempo y los recursos, sino que también mejorará la precisión y la eficiencia en la gestión del conocimiento organizacional y el acceso a la información para la comunidad universitaria, facilitando el aprendizaje, la investigación y los descubrimientos en el campo de la geología.


El contenido de este documento se estructura de la siguiente manera: una visión general del proyecto es presentada en la Sección \ref{sec:intro}. La Sección \ref{sec:sota} explora los trabajos relacionados más relevantes. La Sección \ref{sec:methodology} describe la metodología utilizada, incluyendo los datos y herramientas, así como la arquitectura del sistema, las pruebas realizadas y los resultados obtenidos. Por último, las conclusiones del trabajo realizado y las posibles líneas futuras de desarrollo se enuncian en la Sección \ref{sec:conclusion}.

\section{Trabajos relacionados}
\label{sec:sota}
Las tecnologías de IA tienen el potencial de automatizar muchas de las tareas relacionadas con la investigación y la educación \cite{1-ref14}.
Una de las aplicaciones más sobresalientes son los \textit{sistemas conversacionales}, cuyo desarrollo ha captado gran atención en los últimos años \cite{ref5}\cite{ref13}\cite{ref23}.
Aunque los términos ``sistema conversacional'' y ``chatbot'' tienden a usarse indistintamente, hay una diferencia importante.
Un sistema conversacional es algo más avanzado, ya que busca mantener un diálogo natural y contextual con el usuario, con la posible integración de herramientas externas (web, bases de datos, APIs, etc.).
La literatura sobre esta temática es vasta y reciente, donde numerosos estudios examinan el uso de estos sistemas en diversas áreas.
Esta sección explora algunos trabajos destacados, analizando los objetivos planteados, los datos y métodos utilizados, los resultados obtenidos y las limitaciones identificadas, con el fin de resaltar las contribuciones del presente trabajo.

En \cite{2-ref31} se presenta Geogalactica, un sistema que usa lenguaje en geociencias. Utiliza técnicas avanzadas de aprendizaje automático para analizar grandes conjuntos de datos geológicos y geofísicos con el modelo LLama-7B con 65 millones de tokens de corpus de textos de geociencia.
Este modelo no solo mejora la precisión en la interpretación de fenómenos geológicos, sino que también ofrece nuevas perspectivas sobre la evolución geológica y la interacción de la Tierra con otros sistemas planetarios.

En \cite{3} se propone un framework que genera automáticamente pares de preguntas-respuestas (QA) largas o factoides para evaluar la calidad de RAG. También puede crear conjuntos de datos que evalúan los niveles de alucinación de los LLMs simulando preguntas sin respuesta.
El framework se aplica en la creación de pares de preguntas y respuestas basadas en más de 1000 folletos sobre procedimientos médicos y administrativos de un hospital. La evaluación de especialistas del hospital confirma que más del 50\% de los pares QA son aplicables. Finalmente, se muestra que el marco se puede utilizar para evaluar el rendimiento de Llama-2-13B ajustado en holandés.

En \cite{4-ref3} se analiza cómo los LLMs modernos, basados en la arquitectura Transformer, procesan textos completos para comprender el contexto y generar respuestas precisas. Destaca el uso de modelos como GPT-4 de OpenAI, con 1.76 billones de parámetros, y Llama 2 de Meta AI, con hasta 79 mil millones de parámetros, que permiten especializaciones avanzadas. Estas tecnologías potencian chatbots como ChatGPT, reconocido por su capacidad para generar respuestas naturales; Claude, orientado al razonamiento ético y tareas complejas; y Bing Chat, que combina GPT-4 con acceso a información actualizada y fuentes verificadas, ampliando sus aplicaciones prácticas.

     
En \cite{5-ref19} mejoran la conversión de preguntas en lenguaje natural a consultas SQL mediante un LLM. Se utiliza el conjunto de datos sql-create-context2, con 78,577 ejemplos de preguntas, declaraciones CREATE TABLE y consultas SQL.
Se evaluaron SQLCoder (ajustado en CodeLlama) y LangChain, destacando ambos por su precisión en la generación de consultas SQL.
Al combinar RAG con LangChain, se mejoró la efectividad del sistema. El módulo de compilación de consultas SQL descompone preguntas, recupera ejemplos relevantes y utiliza LLM para generar consultas SQL. Se realizaron experimentos con 100 preguntas en nueve alternativas, utilizando diferentes modelos (GPT-4-32K, GPT-4o, Llama 3.1-405B-Instruct, Mistral Large y Claude 3.5-Sonnet) en plataformas OpenAI y AWS Bedrock, variando en tamaño y capacidad de contexto desde 32K hasta 200K tokens.

Los estudios revisados muestran que la implementación de sistemas conversacionales mejora significativamente la eficiencia de funciones educativas y administrativas.
Un enfoque innovador en este contexto es \textit{Geolog-IA}, un sistema conversacional diseñado para responder tanto a preguntas cualitativas como cuantitativas dentro del ámbito geológico.
Esto permite a usuarios de distintos niveles acceder fácilmente a información relevante, optimizando el aprendizaje y la toma de decisiones en el campo de la geología.


\section{Metodología}
\label{sec:methodology}

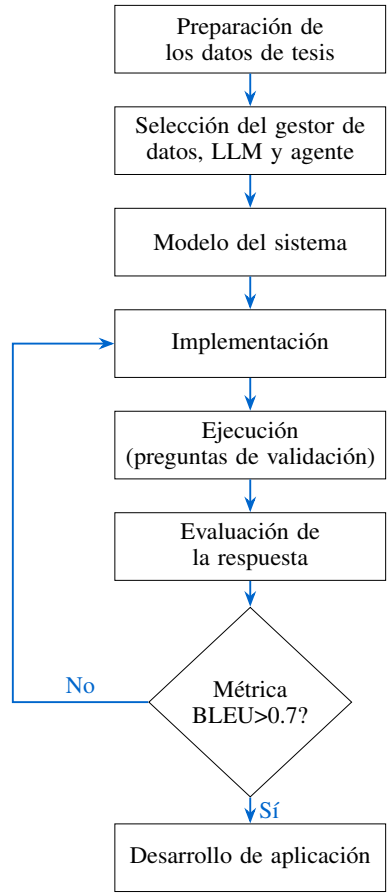
\begin{figure}[!htb]
    \centering
\scalebox{0.9}{    
\begin{tikzpicture}[node distance=1.5cm]
    \node (software) [process] {Preparación de los datos de tesis};
    \node (runproject) [process, below of=software] {Selección del gestor de datos, LLM y agente};
    \node (errors) [process, below of=runproject] {Modelo del sistema};
    \node (CA) [process, below of=errors] {Implementación};
    \node (results) [process, below of=CA] {Ejecución \\ (preguntas de validación)};
    \node (AM) [process, below of=results] {Evaluación de la respuesta};
    \node (studyarea) [decision, below of=AM,yshift=-0.8cm, align=center]
    {Métrica\\ BLEU>0.7?};
    \node (software1) [process, below of=studyarea, yshift=-0.8cm] {Desarrollo de aplicación};

    \draw [arrow] (software) -- (runproject);
    \draw [arrow] (runproject) -- (errors);
    \draw [arrow] (errors) -- (CA);
    \draw [arrow] (CA) -- (results);
    \draw [arrow] (results) -- (AM);
    \draw [arrow] (AM) -- (studyarea);
    \draw [arrow] (studyarea) -- (software1) node[midway,right] {Sí};
    \draw [arrow] (studyarea.west) -- ++(-2,0) node[midway, above] {No} 
    -- ++(0,5.3) -- (CA.west);

    \end{tikzpicture}
}  
    \caption{Metodología utilizada para el desarrollo del sistema conversacional Geolog-IA.}
    \label{fig:flujograma}
\end{figure}

El desarrollo de Geolog-IA sigue el flujo de trabajo presentado en la Figura \ref{fig:flujograma}.
Se propone una metodología que combina un LLM con RAG, una técnica avanzada de recuperación de información para potenciar la precisión de las respuestas.
Para este propósito, se emplea una base de datos como SQLite para estructurar y gestionar la información eficientemente, con el apoyo de librerías especializadas como LangChain para automatizar la creación de prompts.
Como resultado, se asegura una interacción más natural y accesible para los usuarios, con el fin de obtener respuestas coherentes, precisas y contextualizadas sin requerir conocimientos avanzados en SQL.
Para evaluar la calidad del sistema, es fundamental diseñar preguntas y establecer respuestas de validación que puedan cubrir el amplio abanico de los usuarios reales, como docentes, estudiantes, directivos y personal administrativo. A continuación, se explica en detalle cada una de las actividades del flujo de trabajo.


\subsection{Preparación de los datos de tesis}

La preparación adecuada del conjunto de datos de las tesis de titulación es crucial para el éxito del proyecto.
Estas tesis se convierten en la fuente de información esencial para responder las consultas del usuario.
Por medio de una comunicación oficial al Sistema Integrado de Bibliotecas (SIB) de la Universidad Central del Ecuador (UCE), se obtuvo un archivo en formato CSV (Comma Separated Values) que incluye los datos de 244 tesis de pregrado, registradas y actualizadas manualmente hasta diciembre de 2024.
El archivo CSV consta originalmente de 56 campos; sin embargo, no todos son relevantes debido a datos faltantes o innecesarios para nuestro propósito.
Tras el proceso de depuración, se seleccionaron 16 campos esenciales, que conforman un nuevo archivo CSV, el cual permite optimizar el acceso a la información, reduce el tiempo de procesamiento y asegura una operación más eficiente del sistema.
La Tabla \ref{tab:campostesis} describe la estructura de campos del archivo CSV final, el cual es la materia prima para el funcionamiento óptimo del sistema conversacional.

\begin{table}[H]  
 \captionsetup{labelfont=bf, textfont=normal, labelsep=colon} 
 \caption{Estructura del archivo CSV que contiene los datos de las tesis de Geología (FIGEMPA-UCE).}
 \label{tab:campostesis}
 \centering
\scalebox{0.9}{ 
 \begin{tabular}{p{3cm} p{1.5cm} p{6.5cm}}
  \toprule
  \textbf{Campo} & \textbf{Tipo} & \textbf{Descripción} \\
  \midrule
  Id& Texto& Identificador único de la tesis  \\
  \midrule
  título & Texto& Título completo  \\
  \midrule
  autor& Texto& Nombres y apellidos de quien escribió la tesis  \\
  \midrule
  tutor& Texto& Nombres y apellidos de quien dirigió la tesis  \\
  \midrule
  temática& Texto& Área de conocimiento de la tesis  \\
  \midrule
  graduate\_title &Texto& Título profesional obtenido  \\
   \midrule
  thesis\_level& Texto &Nivel académico de la tesis  \\
   \midrule
  carrera &Texto& Carrera o programa académico \\
   \midrule
 year\_approval& Entero &Año en el que se aprobó la tesis  \\
   \midrule
 month\_approval &Entero& Mes en el que se aprobó la tesis  \\
   \midrule
 number\_pages& Entero& Número total de páginas de la tesis  \\
   \midrule
 resumen &Texto& Texto breve que resume el contenido de la tesis \\
   \midrule
  keywords &Texto& Palabras clave relevantes asociadas con la tesis  \\
   \midrule
  citation& Texto &Formato de cita en APA (7ma. Edición) \\
   \midrule
  location &Texto &Ubicación física de la tesis  \\
   \midrule
  url& Texto& Enlace a la versión digital de la tesis \\
  \bottomrule
 \end{tabular}
 }
\end{table}

Un ejemplo completo de un registro del archivo CSV con los campos listados anteriormente se presenta a continuación:

\begin{quote}
\emph{\small
``288b197f-46d3-4483-8698-9fb44c7239ab'',
``Sedimentología y estratigrafía secuencial de la Formación Hollín en el campo Palo Azul - Bloque 18 de la Cuenca Oriente'',
``Yépez Ruiz Andrea Jadira'',
``Zura Quilumbango Cristian Bayardo'',
``Geofísica petrolera'',
``Ingeniería en Geología'',
``Pregrado'',
``Carrera de Ingeniería en Geología'',
2020,
``-'',
132,
``La presente investigación detalla la sedimentología y …'',
``Hollín, Ambientes sedimentarios, Cortejos sedimentarios, Litofacies'',
``Yépez Ruiz, A. (2020) ...'',
``Biblioteca General - FIGEMPA'',
``https://www.dspace.uce.edu.ec/handle/25000/22130''
}
\end{quote}

Cabe notar que el campo ``temática'' fue añadido tras una serie de pruebas preliminares, con el objetivo de categorizar y organizar de manera más eficiente los diferentes temas tratados en las tesis de geología.
Esto facilita la búsqueda y consulta de temas específicos relacionados con las tesis.
Asimismo, permite lograr una recuperación más precisa de la tesis y mejorar las respuestas proporcionadas por el sistema.
Por otra parte, el campo de resumen originalmente se denominaba ``abstract''; sin embargo, las pruebas realizadas determinaron que el cambio de nombre mejoró la localización y extracción de la información.
Este campo es fundamental para ofrecer respuestas coherentes y contextualizadas a los usuarios. Su valor reside en su extensión y riqueza textual, que permite capturar la esencia de la investigación sin la necesidad de integrar el documento completo de la tesis.
Incorporar tesis enteras sería poco eficiente. En su lugar, el resumen provee suficiente información para comprender el tema central, los objetivos, la metodología y los resultados clave. Si el usuario necesita profundizar, se proporciona el enlace URL de la tesis completa, permitiéndole acceder a la fuente original y explorar el contenido en detalle.
Por último, se completó la información ausente con un guion (-) para mantener la uniformidad y facilitar su procesamiento y análisis.

\subsection{Selección del gestor de datos, LLM y agente}

Una vez que se ha preparado convenientemente el conjunto de datos, en esta sección se detallan las herramientas empleadas para su procesamiento.
Puesto que el objetivo es responder preguntas en lenguaje natural usando estos datos, la solución propuesta se basa en un entorno RAG-SQL (generación aumentada por recuperación aplicada a SQL) \cite{ref18}.
Este enfoque se compone de tres elementos principales: el gestor de base de datos que organiza y provee la información estructurada mediante consultas SQL, el LLM que traduce preguntas en lenguaje natural a consultas SQL y genera respuestas precisas y contextualizadas, y el agente que coordina ambos componentes, gestionando el flujo de interacción para entregar al usuario la mejor respuesta posible.

\subsubsection{Gestor de base de datos}
En el diseño de un sistema RAG, el formato de la fuente de información es crucial, ya que impacta directamente en la capacidad del sistema para generar respuestas coherentes, contextualizadas y completas.
En esencia, se tienen dos tipos de fuentes \cite{ref34}: i) \textit{información no estructurada}, que no tiene un modelo de datos predefinido ni una organización específica, como documentos de texto libre, PDFs, imágenes, audio, etc.; y ii) \textit{información estructurada}, que son datos organizados en un formato predefinido, como bases de datos SQL, hojas de cálculo, archivos CSV o JSON, donde la información se almacena en campos y registros con tipos de datos específicos.

Inicialmente, se optó por la versión no estructurada utilizando un archivo PDF que contenía los resúmenes de las tesis. La lógica detrás de esta elección es la simplicidad, pues se trata de texto continuo que puede ser procesado por un LLM para extraer información relevante.
Sin embargo, esta aproximación reveló una limitación significativa: el sistema solo era capaz de ofrecer respuestas a preguntas de carácter cualitativo. Esto significa que podía responder a preguntas como ¿De qué trata esta tesis? o ¿Cuál es la metodología principal mencionada?, ya que estas respuestas se derivan directamente del texto descriptivo del resumen.
Al intentar responder a preguntas que requerían datos específicos o comparaciones numéricas; por ejemplo, ¿Cuántas tesis se publicaron en 2023? o ¿Cuál es el promedio de número de páginas de las tesis? eran difíciles o imposibles de responder con precisión, ya que el sistema no tenía una forma estructurada de acceder y procesar esos datos.

Por ende, se consideró conveniente una fuente de información estructurada para lograr un sistema capaz de responder preguntas de tipo cualitativo y cuantitativo sobre tesis individuales o en su totalidad.
La decisión se vio fuertemente influenciada por la naturaleza del archivo de datos de tesis disponible, que ya venía por defecto en formato CSV, lo cual es intrínsecamente estructurado.
La gran ventaja de esta estructura es que permite al sistema RAG no solo acceder al contenido textual detallado (para respuestas cualitativas), sino también consultar y manipular datos específicos y numéricos (para respuestas cuantitativas).
De esta forma, el sistema puede generar respuestas cualitativas extrayendo información del resumen o de los campos de texto para describir el contenido de una tesis, así como generar respuestas cuantitativas filtrando por año, contando el número de tesis sobre un tema específico o incluso realizando cálculos si los datos numéricos lo permiten.

Por tanto, en lugar de buscar documentos, se consultan tablas en una base de datos relacional \cite{ref26}.
La selección de un adecuado sistema gestor de bases de datos (DBMS) y el diseño de una base de datos de tesis optimizada según las necesidades y condiciones del proyecto son actividades cruciales para la recuperación eficiente de la información y el buen desempeño del sistema conversacional.
Un DBMS proporciona herramientas avanzadas para la definición, manipulación y control de datos, facilitando el desarrollo de aplicaciones y sistemas informáticos \cite{6-ref16}.
Este sistema hace posible la interacción con las tesis a través del lenguaje SQL.
Aquí se consideraron dos de las opciones más populares: SQLite y PostgreSQL, cuyas principales características se despliegan en la Tabla \ref{tab:dbms}.

\begin{table}[H]  
 \captionsetup{labelfont=bf, textfont=normal, labelsep=colon} 
 \caption{Características de los gestores de base de datos utilizados.}
 \label{tab:dbms}

 \centering
 \scalebox{0.9}{

 \begin{tabular}{p{2.5cm} p{6.25cm} p{6.25cm}}
  \toprule
  \textbf{Característica} & 
  \textbf{SQLite} & 
  \textbf{PostgreSQL} \\
  \midrule
  Arquitectura & Sin servidor, integrado & Cliente/Servidor\\
   \midrule
 Tipos de datos & Básicos &Básicos y enriquecidos\\ 
   \midrule
Uso &Aplicaciones integradas, pruebas &Sistemas empresariales, análisis\\
   \midrule
Concurrencia &Escritura limitada (escritor único) & Lectura/Escritura múltiple y simultánea\\
   \midrule
Almacenamiento& Base de datos en un archivo (.sqlite) &Utiliza múltiples archivos gestionados\\
   \midrule
Actuación& Rápido en lecturas simples y monousuario &Optimiza consultas complejas y concurrentes\\
   \midrule
Recursos& Mínimo y ocupa poco espacio &Demanda más memoria y recursos de CPU\\
  \midrule
Licencia&  Dominio público (gratuita)&  Código abierto (licencia PostgreSQL)\\
  \midrule
Acceso& Cargas de trabajo ligeras&  Consultas complejas\\
  \midrule
Limitaciones&No soporta alta concurrencia, rendimiento bajo en grandes volúmenes de datos& Requiere más configuración y recursos, mayor complejidad en la administración\\
  \bottomrule
 \end{tabular}
 }
\end{table}


Inicialmente se utilizó de manera local PostgreSQL por sus mayores capacidades; sin embargo, su uso en la nube es costoso, requiere configuraciones adicionales y permisos de conexión.
SQLite demostró ser una opción más adecuada para satisfacer las exigencias del proyecto.
Es una versión liviana y embebida, ocupa poco espacio y maneja con rapidez la lectura de datos en formato de texto y numérico.
Esta tecnología permite el almacenamiento y acceso a los datos del archivo CSV de una manera estructurada en una o más tablas.
En este caso fue suficiente una sola, debido a que todos los campos del archivo CSV son atributos de una entidad; es decir, cada una de las tesis. 

\subsubsection{Selección del LLM}
\label{subsec:llm}
De similar importancia, es escoger el LLM más conveniente en términos de disponibilidad y rendimiento. Es el componente que se encarga de interpretar las preguntas del usuario en lenguaje natural, traducirlas en lenguaje SQL para recuperar información relevante de la base de datos y generar respuestas coherentes.
Existen múltiples opciones de LLM disponibles en el mercado, cada uno con sus ventajas y desventajas.
En la Tabla \ref{tab:llms} se presentan los modelos comparados, destacando sus principales características.

\begin{table}[H]  
 \captionsetup{labelfont=bf, textfont=normal, labelsep=colon} 
 \caption{Características de los LLMs considerados.}
 \label{tab:llms}
 \centering
 \scalebox{0.83}{
 \begin{tabular}{p{1.5cm} p{2.5cm} p{2.5cm} p{2.5cm} p{2.5cm} p{2.5cm} p{2.5cm}}
  \toprule
  \textbf {Criterio} & \textbf{ChatGPT-4} & \textbf{Claude} & \textbf{Mistral} & \textbf{Llama 3} & \textbf{Llama 3.1} & \textbf{Llama 3.3}\\
  \midrule
 Creador & OpenAI & Anthropic & Mistral AI & Meta & Meta	& Meta \\

  \midrule
 Tipo & Comercial	& Comercial & Gratuito & Gratuito & Gratuito	& Gratuito \\
  \midrule 
  Código & Cerrado &	Cerrado &	Abierto & Abierto	& Abierto	& Abierto \\
 \midrule
 Parámetros & 1T+ & 100B-1T	& 7B-13B	& 8B-70B & 8B-70B	& 8B-70B\\
  \midrule
 Uso & Chat e IA general	& Chat y análisis de texto	& Conversación e inferencia rápida	& General y conversación	& Optimizado para chat	& Conversación respuesta lenta\\
   \midrule
 DBMS & Requiere API	& Requiere API	& Sin API & Sin API	 & Sin API & Sin API \\
 \midrule
 Calidad  & Excelente & Muy buena & Muy buena & Muy buena & Muy buena & Muy buena \\
 \midrule
Respuesta & Precisa, coherente y rápida & Razonable y coherente & Razonable y coherente & Óptima, razonable y coherente & Rápida y razonable & Lenta y coherente \\
 \midrule
 Fecha	& Nov 2024	& Ago 2024 & Dic 2023 & Jun 2023 & Nov 2024	& Dic 2024 \\
 \midrule
 Soporte & Continuo & Limitado & Limitado & Activo & Continuo & Limitado\\
  \bottomrule
 \end{tabular}
 }
\end{table}

Los factores que influyeron para la elección fueron el costo, la accesibilidad, la capacidad de integración con bases de datos y el rendimiento.
Se descartaron GPT-4 y Claude por no ser gratuitos, a pesar de ofrecer integración directa con bases de datos y alto rendimiento en generación de texto.
Aunque son modelos altamente optimizados, su acceso restringido y licencias de pago los hacen inviables para el proyecto.
Dentro de Llama, la versión 3 fue descartada por ser menos optimizada, mientras que Llama 3.3, a pesar de ser la versión más reciente, no cuenta con suficiente documentación ni soporte continuo, lo que limita su fiabilidad.
Llama 3.1 fue seleccionado debido a su equilibrio entre estabilidad, eficiencia y optimización en la generación de respuestas, con soporte activo y actualizaciones recientes, lo que garantiza mayor confiabilidad y rendimiento en el proyecto.

\subsubsection{Agente SQL}
La falta de este agente ocasionaría dos problemas fundamentales: i) el usuario debería dominar el lenguaje SQL y la estructura interna de la base de datos para consultar y obtener la información de su interés; y ii) la pregunta del usuario pasaría directamente al LLM, el cual respondería basado únicamente en la información con la que fue entrenado, lo que puede llevar a respuestas genéricas, alucinaciones y carentes de contexto.

Por ende, el agente es un componente central del sistema propuesto, ya que actúa como un intermediario que aprovecha el potente razonamiento del LLM a través de prompts especializados.
Así, es capaz de guiar al LLM para interpretar y convertir una pregunta en lenguaje natural
en consultas SQL y recuperar información relevante, así como en la generación de respuestas coherentes, precisas y contextualizadas.

En la práctica, no es necesario implementar todas estas funcionalidades desde cero. Se ha optado por utilizar el framework de código abierto \textit{LangChain} \cite{7-ref15}.
Su nombre combina “Lang” (lenguaje) y “Chain” (cadena), reflejando su arquitectura modular y capacidad para conectar múltiples componentes y estructurar flujos de trabajo eficientes,
facilitando el desarrollo de aplicaciones avanzadas basadas en IA.

LangChain ofrece agentes SQL listos para usar, se seleccionó un agente del tipo \textit{ZERO\_SHOT\_REACT\_DESCRIPTION}, el cual combina tres pilares clave: flexibilidad, ya que puede tomar decisiones y resolver una tarea sin ejemplos o interacciones previas (Zero-Shot); un mecanismo iterativo de razonamiento y acción (ReAct): $Thought \rightarrow Action \rightarrow Observation$, que le permite planificar, ejecutar y corregir errores; y una descripción clara y precisa de cada herramienta disponible que le permita decidir la más adecuada para avanzar hacia la solución \cite{ref24}\cite{ref30}.




\subsection{Modelo del sistema}
Esta sección describe la solución propuesta y explica cómo interactúan los componentes del sistema conversacional.
Primero, se analiza el funcionamiento del LLM de manera particular, ya que constituye el elemento central para la comprensión de preguntas y generación de respuestas.
Seguidamente, se presenta el esquema y funcionamiento del sistema RAG-SQL de manera integral, abarcando todos sus elementos y el flujo general del procesamiento de consultas.

\subsubsection{Modelo de lenguaje extenso (LLM)}
Actualmente, un LLM puede ser considerado como el avance más significativo dentro del campo del procesamiento del lenguaje natural (NLP), ya que hace posible que la máquina pueda entender, interpretar y generar lenguaje natural como el ser humano \cite{ref35}.
En este caso, constituye el núcleo del sistema conversacional, siendo el ``cerebro'' detrás de su funcionamiento.
Por esta razón, entender su operación es fundamental para comprender las capacidades del sistema.

\begin{figure}[!htb]
    \centering

\begin{tikzpicture}[
    font=\small,
    box/.style={rectangle, draw=blue!50, thick, fill=blue!10, minimum width=5.5cm, minimum height=1cm, align=center},
    arrow/.style={-{Stealth}, thick},
    node distance=0.75cm
]

\node (input) at (0, 0) {\textbf{Prompt:} \texttt{``El clima es''}};

\node[box, below=of input] (token) {Tokenización y Embedding\\(subpalabras → vectores)};
\node[box, below=of token] (posenc) {Codificación Posicional\\(suma con embeddings → orden)};
\node[box, below=of posenc] (attention) {Mecanismo de Atención\\(Masked Self-Attention)};
\node[box, below=of attention] (ffn) {Red Feedforward\\(no linealidad punto a punto)};
\node[box, below=of ffn] (softmax) {Softmax Final\\(distribución de probabilidad)};

\node[below=0.75cm of softmax] (output) {\textbf{Salida:} \texttt{``soleado''}};

\draw[arrow] (input) -- (token);
\draw[arrow] (token) -- (posenc);
\draw[arrow] (posenc) -- (attention);
\draw[arrow] (attention) -- (ffn);
\draw[arrow] (ffn) -- (softmax);
\draw[arrow] (softmax) -- (output);

\end{tikzpicture}

    \caption{Esquema de un transformer sólo decodificador.}
    \label{fig:transformer}
\end{figure}
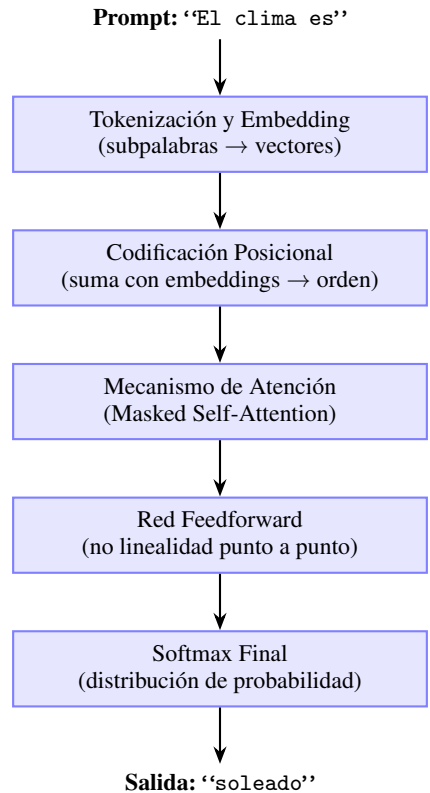

Un LLM es un sistema de IA basado en redes neuronales artificiales; en particular, una arquitectura denominada \textit{Transformer sólo decodificador} \cite{8-ref4}. Para explicar su funcionamiento, es posible organizar su estructura en cinco niveles, de arriba hacia abajo, tal como se muestra en la Figura \ref{fig:transformer}.

En el nivel 1, se ingresa el mensaje (\textit{prompt}) al LLM. La preparación conveniente del texto para su posterior procesamiento requiere la división de la frase de entrada en palabras o subpalabras como unidades básicas de procesamiento (\textit{tokens}); cada una se convierte en una representación numérica vectorial (\textit{embedding}), que tiene un significado semántico propio de cada token.

El nivel 2 corresponde a una \textit{codificación posicional}. Puesto que todas las palabras entran al mismo tiempo, un mecanismo de orden debe introducir la posición de la palabra dentro de la secuencia.
Dado que los transformers no procesan el texto de manera secuencial como las redes recurrentes, este proceso es fundamental para que el modelo entienda la estructura del lenguaje y las relaciones entre términos según su posición.

En el nivel 3, el bloque de atención es clave para la comprensión del texto.
A través de un mecanismo de \textit{auto-atención} es posible determinar las relaciones entre diferentes palabras para añadir contexto a cada palabra y entender su significado  dentro de la frase.
Se trata de un problema de búsqueda de dependencias semánticas, donde cada palabra identifica las palabras que más influyen en su significado. Se definen los siguientes pasos:

\begin{itemize}
    \item A partir de cada token de entrada, se crean tres versiones: Query (Q), Key (K) y Value (V), los cuales interactúan entre sí.
    \item La consulta Q se compara con cada clave K para calcular pesos de atención.
    \item Los pesos se aplican a los vectores de valor V, que luego se combinan para generar una representación contextual enriquecida de cada token de entrada.
\end{itemize}
 
Como resultado, se obtienen vectores con significado semántico enriquecido con contexto; es decir, aquí se realiza un ajuste del significado de cada palabra según el contexto de la frase.
La salida del mecanismo de atención pasa al nivel 4, donde una red neuronal \textit{feed-forward} aplica transformaciones lineales con parámetros entrenables y funciones de activación no lineales.
A diferencia de la atención, esta red no modela relaciones entre palabras, sino que actúa sobre cada token de manera independiente, permitiendo refinar y enriquecer las representaciones ya contextualizadas de cada palabra.

Finalmente, en el nivel 5 se aplica una transformación lineal para mapear cada vector a uno de dimensión igual al espacio de salida (\textit{logits}), el cual es convertido en una distribución de probabilidad sobre las palabras del vocabulario mediante la función \textit{softmax}.
Solamente el último vector es considerado para la predicción, que consiste en seleccionar la palabra con mayor puntuación y generarla como salida.

Los pasos descritos son realizados de manera secuencial y autoregresiva; es decir, una vez que se predice el primer token de la respuesta, este se añade al final de la secuencia de entrada, y el proceso se repite para predecir el siguiente token.
Este diseño y funcionamiento modular permite procesar una secuencia de texto de entrada de manera eficiente y generar texto de salida con un alto grado de precisión y coherencia \cite{9-ref28}.

Este tipo de modelos es entrenado con una cantidad masiva de datos (corpus) en dos etapas: i) el \textit{preentrenamiento} para aprender la estructura del lenguaje y predecir el siguiente token de una secuencia; y ii) el \textit{postentrenamiento} o afinamiento de tipo supervisado para que el modelo siga instrucciones (prompts) \cite{ref7}, responda preguntas y adquiera otras habilidades.   

Existen múltiples LLMs desarrollados por diferentes organizaciones, cada uno con fortalezas y limitaciones, como los analizados en la Sección \ref{subsec:llm}. Se ha optado por Llama 3.1, creado por Meta, el cual es un modelo de código abierto optimizado para el razonamiento y la conversación, ampliamente accesible para investigación y aplicaciones prácticas \cite{ref27}. A pesar de sus sorprendentes capacidades, su uso puede presentar inconvenientes como alucinaciones y desactualización de conocimiento, algo común en los LLMs actuales.
Para enfrentar estas limitaciones, los documentos de tesis podrían servir como datos de entrenamiento para llevar a cabo un \textit{fine-tuning} o ajuste del modelo; sin embargo, este proceso es complejo y costoso. Como alternativa, se implementa la estrategia de RAG basada en SQL.

\subsubsection{Arquitectura RAG-SQL}

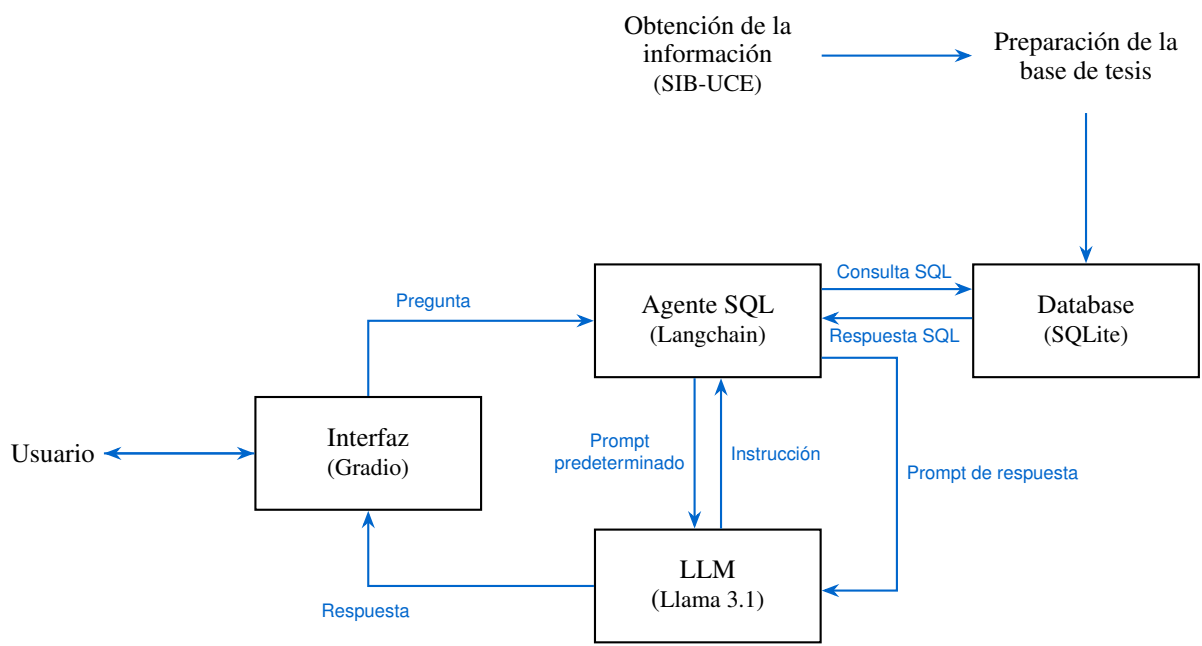
\begin{figure}[H]
    \centering
    \begin{tikzpicture}[node distance=2cm]
		
		\node[box] (sqlagent) {Agente SQL\\{\small (Langchain)}};
		\node[box, below=of sqlagent] (llm) {LLM\\(\small Llama 3.1)};
		
		\node[box, left=3cm of $(sqlagent)!0.5!(llm)$] (interface) {Interfaz \\ \small (Gradio)};
		
		\node[left=of interface] (user) {Usuario};
		
		\node[box, right=of sqlagent] (db) {Database\\{\small (SQLite)}};
		
		\node[box, draw=none, above=of db] (prep) {Preparación de la \\ base de tesis};
		\node[box, draw=none, left=of prep] (info) {Obtención de la \\ información \\ \small (SIB-UCE)};
		
		\draw[arrow] (user) -- node[above, label] {} (interface);
		\draw[arrow] (interface) -- node[above, label] {} (user);
		
		\draw[arrow] (interface.north) -- ++(0,1) -- node[above, label, pos=0.3] {Pregunta} (sqlagent.west);
		
		\draw[arrow] (llm.west) -- ++(-3,0) -- node[above, label, pos=-0.6] {Respuesta} (interface.south);
		
		\draw[arrow] ([yshift=12pt]sqlagent.east) -- node[above, label, pos=0.5] {Consulta SQL} node[below, label, pos=0.5] {} ([yshift=12pt]db.west);

		\draw[arrow] ([yshift=1pt]db.west) -- node[below, label, pos=0.5] {Respuesta SQL} node[below, label, pos=0.5] {} ([yshift=1pt]sqlagent.east);
				
		
		\draw[arrow] ([xshift=-5pt]sqlagent.south) -- node[left, label] {Prompt\\predeterminado} ([xshift=-5pt]llm.north);

		\draw[arrow] ([xshift=5pt]llm.north) -- node[right, label] {Instrucción} ([xshift=5pt]sqlagent.south);

		\draw[arrow] ([yshift=-14pt]sqlagent.east) -- ++(1,0) |- 
		node[right, label, pos=0.25] {Prompt de respuesta} 
		([yshift=20pt]llm.south east);
		
		\draw[arrow] (info) -- (prep);
		\draw[arrow] (prep) -- (db);
		
		
	\end{tikzpicture}
    \caption{Arquitectura del sistema conversacional de tesis: Geolog-IA.}
    \label{fig:architecture}
\end{figure}

RAG es una técnica que combina un LLM con un mecanismo de recuperación de información precisa y actualizada.
En este caso, la información se encuentra en una base de datos SQL, la cual debe ser accedida y consultada.
Sin embargo, el modelo LLM por sí solo no puede ejecutar acciones; se necesita el agente para interactuar con la base de datos, lo que permite la generación y validación de consultas SQL.
Por ende, el agente es quien coordina el flujo de información del sistema conversacional, cuya arquitectura se esquematiza en la Figura \ref{fig:architecture}. 




El funcionamiento general del sistema puede describirse en los siguientes pasos:

\begin{enumerate}
    \item \textbf{Pregunta del usuario}: el proceso comienza cuando el usuario ingresa una pregunta en lenguaje natural relacionada con las tesis, por ejemplo:
    \begin{center}
    \emph{¿Cuántas tesis se realizaron en 2022?}        
    \end{center}
    Para facilitar el uso del sistema, se ha implementado una interfaz gráfica de chat desarrollada con el framework \textit{Gradio} que actúa como el punto de entrada del sistema.

    \item \textbf{El agente recibe la pregunta y construye el prompt}: el agente captura la pregunta del usuario y la integra en su prompt predeterminado, el cual guía al LLM para la traducción a código SQL.
    Este prompt especifica el rol del agente, la descripción de las herramientas disponibles, el manejo de errores, el mecanismo iterativo de razonamiento y acción ($ Thought \rightarrow Action \rightarrow Action Input \rightarrow Observation $) y la pregunta del usuario. A continuación, un ejemplo omitiendo información secundaria:

\begin{quote}
\small{
``You are an agent designed to interact with a SQL database. Given an input question, create a syntactically correct sqlite query to run, then look at the results of the query and return the answer.\\
...
\\
You have access to tools for interacting with the database. Only use the below tools. Only use the information returned by the below tools to construct your final answer.\\
...

You must double check your query before executing it. If you get an error while executing a query, rewrite the query and try again.\\
...
\\
sql\_db\_query - Input to this tool is a detailed and correct SQL query, output is a result from the database
sql\_db\_schema - Input to this tool is a comma-separated list of tables, output is the schema and sample rows ...\\
sql\_db\_list\_tables - Input is an empty string, output is a comma-separated list of tables in the database.\\
sql\_db\_query\_checker - Use this tool to double check if your query is correct before executing it.\\
...\\
Use the following format:

Question: the input question you must answer\\
Thought: you should always think about what to do\\
Action: the action to take, should be one of [{tool\_names}]\\
Action Input: the input to the action\\
Observation: the result of the action\\
... (this Thought/Action/Action Input/Observation can repeat N times)\\
Thought: I now know the final answer\\
Final Answer: the final answer to the original input question\\

Begin!

Question: ¿Cuántas tesis se realizaron en 2022?\\
Thought: I should look at the tables in the database to see what I can query.  Then I should query the schema of the most relevant tables.
''
}
\end{quote}

    \item \textbf{El LLM interpreta el prompt}: 
    el modelo sigue la indicación de que debe convertir la pregunta en lenguaje natural a una instrucción SQL que pueda ser procesada por la base de datos. Aquí se genera un texto que podría estar mezclado con otras explicaciones o formato no estructurado.

    \item \textbf{El agente recibe la salida del LLM}: el agente debe extraer y depurar la consulta SQL generada, asegurando que esté lista para su ejecución. Por ejemplo, para la pregunta ¿Cuántas tesis se realizaron en 2022?, se obtiene como salida:
    
    \begin{center}
    \emph{SELECT COUNT (*) FROM tesis WHERE Año\_Aprobación = 2022;}
    \end{center}

    



    \item \textbf{Ejecución de la consulta SQL}:
    el agente usa la herramienta respectiva para ejecutar el SQL generado. SQLite procesa la consulta y devuelve el resultado al agente. En este caso,  si la base de datos contiene 10 registros de tesis aprobadas en 2022, el resultado obtenido será el número 10.

    \item \textbf{Generación de la respuesta final}:   
    el agente utiliza un nuevo prompt que será enviado al LLM.
    Este prompt incluye la pregunta original del usuario, la consulta SQL ejecutada y el resultado devuelto por la base de datos.

\begin{quote}
\small    ``Dada la siguiente pregunta del usuario sobre las tesis de Geología y el resultado SQL, genera una respuesta detallada al usuario mencionando siempre las tesis de Geología como tópico principal.\\
Question: \{question\}\\
SQL Result: \{result\}''
\end{quote}
   
    Este es el contexto que interpreta el LLM para producir una explicación natural y coherente al usuario. Por ejemplo, la respuesta final podría ser:
    \begin{center}
    \emph{En el año 2022, se realizaron 10 tesis en la carrera de Ingeniería en Geología de la FIGEMPA.}    
    \end{center}
       
\end{enumerate}

Finalmente, la respuesta generada es enviada a la interfaz de Gradio, donde se muestra al usuario junto con la pregunta original y la consulta SQL utilizada para obtener el resultado.
El flujo descrito garantiza una interacción eficiente entre el usuario, el agente, la base de datos SQLite y el LLM Llama 3.1, permitiendo la conversión automática de consultas en lenguaje natural a SQL y la generación de respuestas precisas y comprensibles.

%

\subsection{Implementación del sistema}
\label{subsec:coding}

Tras la descripción detallada de la arquitectura del sistema RAG-SQL,
se procede a su implementación y posterior ejecución.
Debido a las altas exigencias computacionales de un LLM moderno, se aprovecha la plataforma \textit{Google Colab} \cite{ref10}, la cual ofrece los recursos técnicos apropiados para el procesamiento en la nube de este tipo de aplicaciones.

Entre las especificaciones técnicas más importantes, se tienen un procesador Intel Xeon de 2.00 GHz, RAM de 13.61 GB, almacenamiento de 120.94 GB y tarjeta gráfica Tesla T4 con RAM de 15.38 GB. El sistema operativo es Linux X86\_64, kernel 6.1.85+, distribución Ubuntu 22.04.4 LTS. El lenguaje de programación Python v3 y se utiliza la librería de soporte Langchain \cite{10-ref11}; en particular, langchain\_ollama y langchain\_community para interactuar con Llama 3.1 \cite{11-ref20} y conectarse con herramientas y servicios proporcionados por la comunidad, respectivamente. Además, la librería colab-xterm facilita un terminal interactivo de línea de comandos en el entorno del cuaderno de programación (\textit{notebook}).

El código del sistema conversacional se estructura en los siguientes módulos:

\begin{itemize}
    \item \textit{Conexión a la BDD}: se encarga de establecer la conexión a la base de datos SQLite (archivo \textit{tesis.db}) almacenada en Google Drive para que la librería LangChain pueda interactuar con ella y realizar consultas.

    \item \textit{Cargar el LLM}: mediante un terminal interactivo, se ejecutan los comandos para la descarga e instalación del servidor de LLMs Ollama, así como la puesta en marcha de Llama 3.1. Cabe señalar que podría ser necesario ejecutar los comandos más de una vez para lograr la inicialización correcta del servidor y el LLM.

    \item \textit{Agente SQL}: aquí se crea y configura un agente inteligente de LangChain capaz de interactuar con la base de datos, a través de un prompt interno que define un modo iterativo de razonamiento y acción para llegar a la respuesta final.
    \item \textit{Módulo RAG}: define el flujo completo con los elementos anteriores.
    Recibe la pregunta en lenguaje natural, utiliza el agente para interactuar con Llama3.1, el cual determina las acciones necesarias, como generar una consulta SQL, que el agente ejecuta en la base de datos para recuperar información (``Retrieval''). Luego, utiliza el LLM nuevamente, junto con la información recuperada, para generar una respuesta coherente en lenguaje natural (``Augmented Generation'') basada en los resultados para el usuario.
\end{itemize}

Cabe resaltar que uno de los parámetros más influyentes en el comportamiento del sistema es la \textit{temperatura}, un parámetro que se establece en el momento de crear la instancia del LLM (en este caso, el valor por defecto 0.5) y que controla el nivel de creatividad y variabilidad en las respuestas: valores más altos generan respuestas más creativas y variadas, mientras que valores más bajos producen respuestas más precisas y coherentes.


  

 



\subsection{Ejecución}

Una vez implementado el sistema RAG-SQL, se lo pone a prueba con preguntas de validación que reflejen fielmente las necesidades y expectativas del público objetivo.
Esto permite asegurar una evaluación más precisa y relevante del sistema conversacional. 
Con el propósito de entender qué tipo de información buscarían los usuarios, se llevó a cabo una encuesta a una muestra de 55 usuarios potenciales.
La Tabla \ref{tab:survey} presenta las preguntas incluidas en dicha encuesta y los resultados obtenidos.



\begin{table}[!htb]
    \captionsetup{labelfont=bf, labelsep=colon} 
    \caption{Preguntas de la encuesta a los usuarios del sistema conversacional y resultados obtenidos.}
    \label{tab:survey}
    \centering
    \begin{tabular}{|c|m{5cm}|m{8cm}|}
    \hline
        \textbf{No}. & \textbf{Pregunta} & \textbf{Resultados} \\
        \hline    
         1 & ¿Cuál es su perfil de usuario? &  \includegraphics[width=0.63\linewidth]{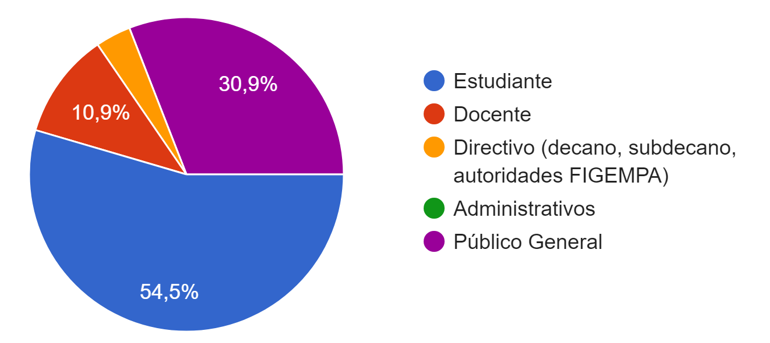} \\
         \hline
         2 & ¿Crees que el desarrollo del sistema conversacional sería útil para consultas sobre tesis de geología? &  \includegraphics[width=0.67\linewidth, height=2cm]{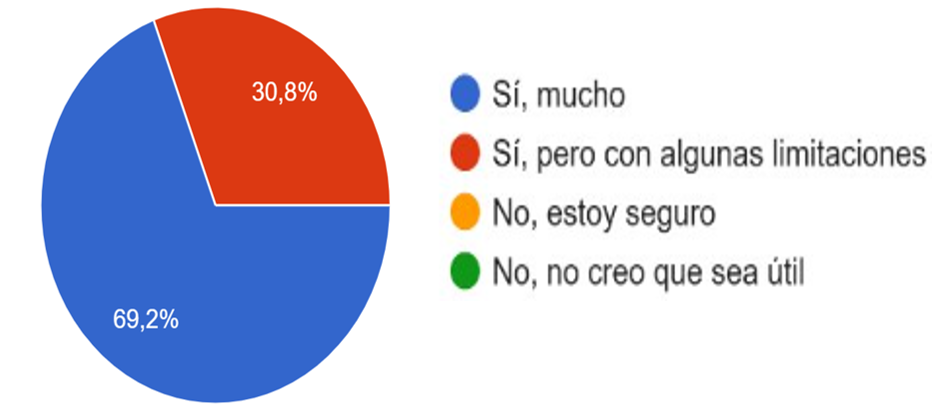} \\
         \hline
         3 & ¿Qué tan útil encontraste la información proporcionada por el sistema conversacional? &  \includegraphics[width=0.53\linewidth]{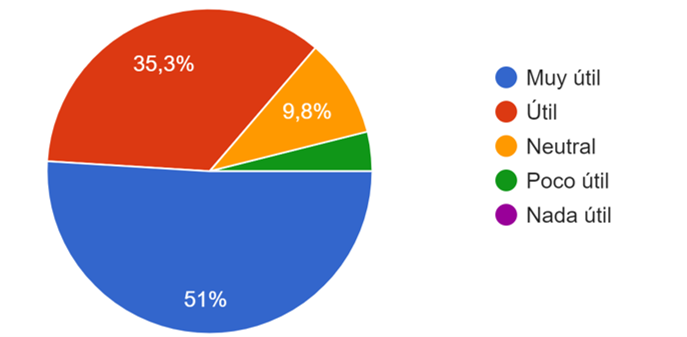} \\
         \hline
         4 & ¿Cómo calificarías la claridad de las respuestas del sistema conversacional? &  \includegraphics[width=0.55\linewidth]{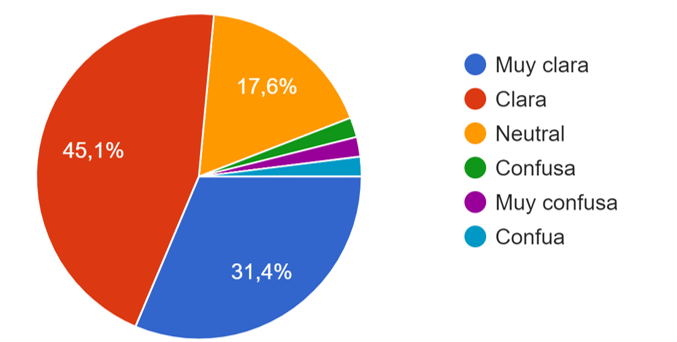} \\
         \hline
         5 & ¿Qué tan rápido recibiste una respuesta del sistema conversacional? &  \includegraphics[width=0.53\linewidth]{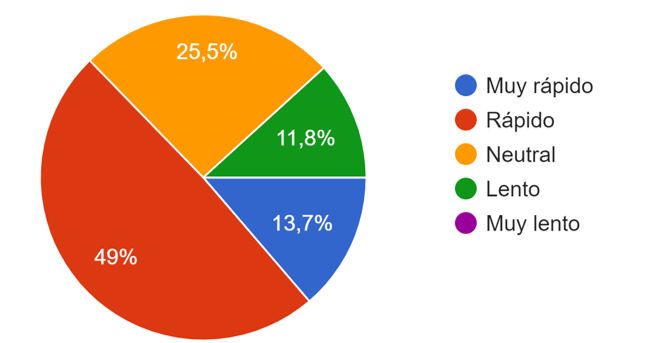} \\
         \hline
         6 & ¿El sistema conversacional entendió correctamente tus preguntas? &  \includegraphics[width=0.64\linewidth]{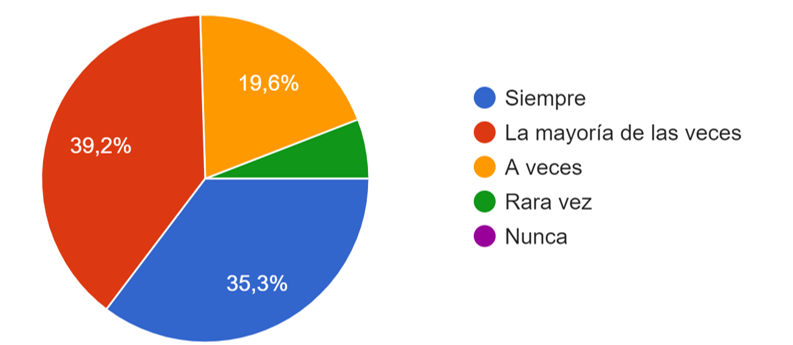} \\
         \hline
         7 & ¿Qué preguntas le harías a un chat especializado en tesis de geología? & Directivos: ¿En qué año se realizaron más tesis?; estudiantes: ¿Qué tutor ha dirigido más tesis?; docentes: ¿Qué temas de tesis se han abordado con mayor frecuencia?; administrativos: ¿Cuáles son los temas y títulos más comunes de las tesis? \\
         \hline
         8 & Respecto a las tesis de geología, ¿qué temas te gustaría conocer? & La mayoría de los encuestados mostró interés en obtener información sobre autores, temas específicos y las diversas temáticas que se han abordado en las tesis.\\
         \hline
    \end{tabular}
\end{table}

La mayoría de los participantes respondió afirmativamente, destacando la gran utilidad del sistema. Sin embargo, algunos que ya habían probado sistemas conversacionales previamente mencionaron que, aunque les parecía útil, habían encontrado ciertas limitaciones en el desempeño de estos sistemas.
En el caso de las preguntas más mencionadas por parte de los diferentes perfiles de usuario, las respuestas fueron diversas; sin embargo, estas preguntas se enfocaron principalmente en temas relacionados con el ámbito académico de la geología.

Con base en esta retroalimentación directa, fue posible plantear un conjunto de preguntas de validación que no solo probaran las capacidades técnicas del sistema conversacional, sino que también garantizaran que fuera realmente útil y relevante para las necesidades del público objetivo. Así, hemos desarrollado una amplia base de preguntas específicas alineadas con las características de cada perfil de usuario. En el enlace:  
\href{https://uceedu-my.sharepoint.com/:f:/g/personal/adguatumillo_uce_edu_ec/ErKQOUhaPw5OpFTQrgYZKkUBoWFwIEs6pkhz1QSADhXjFg?e=PxuEbd}{OneDrive}, se encontrará un archivo de Excel con una lista detallada de estas preguntas, así como capturas de pantalla de las preguntas realizadas al sistema.
La Tabla \ref{tab:table1e} presenta una muestra de estas preguntas de validación con ejemplos representativos de cada perfil de usuario. También se incluyen las respuestas tanto del sistema conversacional como aquellas almacenadas en la base de datos. Estas respuestas son comparadas en la siguiente etapa.

\begin{table}[!htb]  
 \captionsetup{labelfont=bf, labelsep=colon} 
\caption{Evaluación de las respuestas sobre las tesis de Geología, FIGEMPA-UCE.}
 \centering
 \small 
\begin{tabular}{>{\centering\arraybackslash}m{1.6cm} m{4 cm}  m{2.7cm}  m{7cm}}   
  \toprule
  \textbf{Usuario} & \textbf{Pregunta} & \textbf{Respuesta SQLite} & \textbf{Respuesta del sistema conversacional}   \\
  \midrule
  \rotatebox{90}{Directivos} & ¿Tutor con más tesis en el año 2022? & Troncoso Salgado Liliana Paulina | Total tesis 6 & ``Según los datos de las tesis en Geología, el tutor con más tesis aprobadas en el año 2022 es Troncoso Salgado Liliana Paulina.''  \\
  \midrule
  \rotatebox{90}{Estudiantes} & ¿Recomienda un tutor para tesis con temática volcanismo? & Bustillos Arequipa Jorge Eduardo 
  
  Ruiz Paspuel Andrés Gorki & ``Si estas buscando un tutor para realizar una tesis relacionada con el tema de volcanismo, la base de datos sugiere varios nombres que podrían ser útiles. Los resultados obtenidos, se recomienda a Andrés Gorki Ruiz Pasquel como posiblle tutor y a Jorge Eduardo Bustillos Arequipa como otra opción.'' \\
\midrule
\rotatebox{90}{Administrativos} & ¿Existe alguna tesis con el título 'Análisis sedimentológico para la generación del modelo estético de la Arenia C inferior del campo Armadillo'? & 1 & ``La consulta ha encontrado un resultado con el título especificado.'' \\
\midrule 
 \rotatebox{90}{Docentes} & ¿Proporcióname el título de la tesis de Carrillo Guerra  Janina Lisbeth? & Determinación del potencial de acumulación hidrocarburífero de las calizas “A” y “M2” de la formación Napo, campo Drago & ``La tesis de
Janina Lisbeth Carrillo Guerra es titula: "Determinación del potencial de acumulación hidrocarburífero de las
calizas 'Y' y 'M2' de la formación Napo, campo Drago". Espero que esta información sea útil para ti en tu
investigación sobre Geología. ¿Necesitas algo más?'' \\
  \bottomrule
 \end{tabular}
 \label{tab:table1e}
\end{table}

La encuesta realizada fue una herramienta invaluable para identificar directamente las necesidades y expectativas de los usuarios.
También, el análisis de las respuestas permitió estructurar mejor la base de datos y agregar información relevante.
Esto asegura optimizar el sistema para que sea más útil y eficiente para los beneficiarios del proyecto, mejorando la experiencia en general.

\subsection{Evaluación con BLEU}

Una vez identificadas las preguntas más representativas, se prueban en el sistema conversacional para verificar su funcionamiento.
La precisión y coherencia de las respuestas generadas son evaluadas con la métrica \textit{BLEU} (Bilingual Evaluation Understudy) \cite{ref6}\cite{ref22}. 
Esta métrica es ampliamente utilizada para evaluar la calidad de textos generados automáticamente en comparación con una referencia humana.
Se basa en la coincidencia lingüística entre el texto generado y el texto de referencia, ponderando entre 0 y 1 la precisión y penalizando en caso de generar textos más largos de lo esperado \cite{ref9}.

En este caso, BLEU se ha adaptado para comparar las respuestas generadas por el sistema conversacional y las respuestas obtenidas mediante consultas SQL a la base de datos.
En este contexto, se han definido criterios de evaluación flexibles, considerando coincidencias parciales de palabras clave y valores numéricos dentro de las respuestas.
En particular, si una respuesta generada contiene al menos un número en común con la esperada, se le asigna un puntaje elevado (hasta 1).
Si hay coincidencias en palabras clave, pero no en números, el puntaje varía progresivamente entre 0.6 y 1, dependiendo del grado de coincidencia léxica. Cuando no haya coincidencias significativas, se aplica BLEU tradicional con un factor de ajuste, limitando la puntuación máxima a 0.4.

Estos ajustes permiten que la evaluación refleje mejor la utilidad práctica de las respuestas del sistema en un contexto técnico, donde la presencia de valores específicos y términos clave es más relevante que una coincidencia exacta en la redacción.
Los resultados obtenidos con esta metodología permiten analizar la efectividad del sistema conversacional y ajustar su desempeño para mejorar la precisión en sus respuestas.

\begin{figure}[!htb]
\centering
\includegraphics[width=\linewidth]{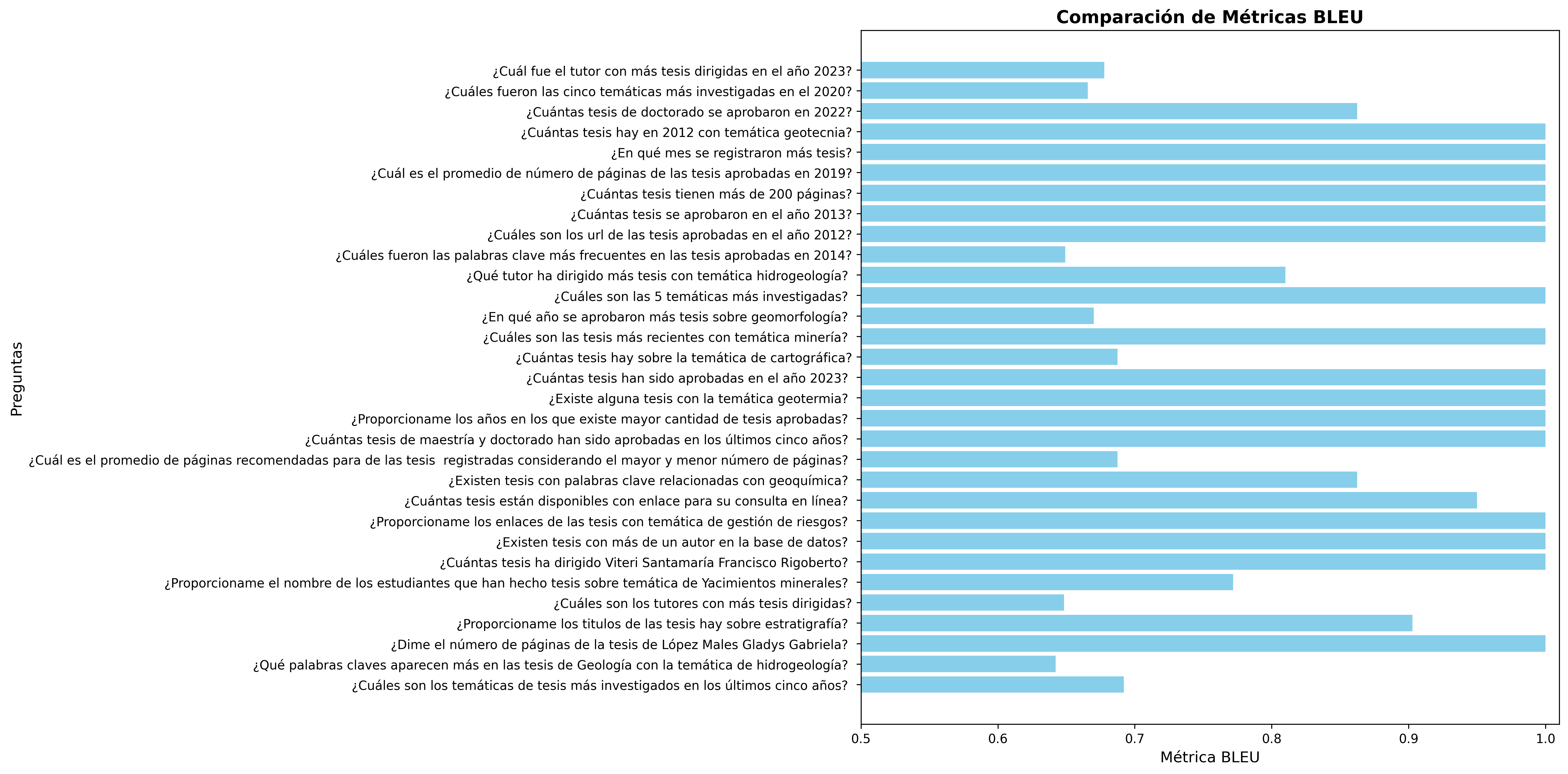}
\caption{Comparación de la métrica BLEU para las respuestas del sistema conversacional.}
\label{fig:blue}
\end{figure}

La Figura \ref{fig:blue} muestra la comparación de los valores de BLEU obtenidos en diferentes consultas realizadas al sistema sobre las tesis de geología.
Se observa que algunas preguntas presentan valores de BLEU cercanos a 1, lo que indica una alta coincidencia entre la respuesta generada y la respuesta esperada.
Esto ocurre en consultas donde los datos numéricos y categóricos parecen haber sido bien interpretados, mientras que hay preguntas con valores más bajos de BLEU, lo que sugiere que el modelo en ocasiones requiere una interpretación más compleja del lenguaje.


Tomando en cuenta los valores del BLEU para las respuestas analizadas, se tiene un promedio de 0.87 y un BLEU $>$ 0.6 suele considerarse aceptable en asistentes conversacionales \cite{ref9}. Entonces, los resultados muestran que el sistema propuesto tiene un buen desempeño en consultas estructuradas tanto cuantitativas como cualitativas.

\subsection{Aplicación web: Geolog-IA}
Nuestro objetivo es proporcionar un sistema conversacional sobre tesis de geología que sea fácil de usar y disponible públicamente.
La solución propuesta recibe el nombre de \textit{Geolog-IA}, dándole identidad propia y que lo distingue de productos similares.
Geolog-IA ha sido implementado mediante Google Colab y Hugging Face Spaces.

\subsubsection{Versión en Google Colab}
Es la implementación original explicada en la Sección \ref{subsec:coding}, que puede ejecutarse en el cuaderno interactivo de Google Colab. Esta versión está orientada al desarrollo y la experimentación, permitiendo a los usuarios la integración con recursos externos como Google Drive, la modificación directa de parámetros y la ejecución del código de manera personalizada. El notebook del proyecto está disponible para la comunidad en \textit{GitHub}, accesible a través del siguiente enlace: \url{https://github.com/cimejia/llm-rag-sql/tree/main}.

\begin{figure}[H]
\centering
\fbox{\includegraphics[width=0.95\linewidth]{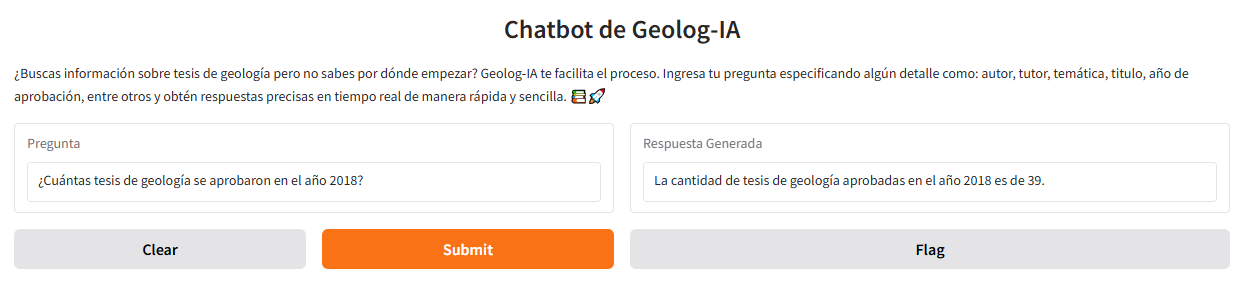}}
\caption{Interfaz gráfica del sistema conversacional creada con Gradio.}
\label{fig:gui}
\end{figure}

Se ha integrado al final del notebook una interfaz gráfica de tipo chat utilizando la librería \textit{Gradio} \cite{12-ref1}.
Esta librería proporciona una solución rápida y eficiente para crear una aplicación web intuitiva y atractiva para los usuarios \cite{13-ref2}.
La interfaz se ha configurado visualmente en dos secciones principales (Figura \ref{fig:gui}).
A la izquierda, el usuario ingresa una pregunta en el cuadro de texto etiquetado como ``Pregunta''.
A la derecha, se muestra el campo de ``Respuesta Generada'', donde aparecerá la salida del sistema.
Además, se incluyen botones como ``Submit'' para procesar la pregunta y ``Flag'' para reportar respuestas incorrectas o inadecuadas.
El usuario puede ingresar su consulta y, tras presionar el botón ``Submit'', el sistema procesa la información y muestra la respuesta correspondiente.
Además, se ha activado la función \textit{allow\_flagging=``manual''} para permitir el reporte de respuestas erróneas mediante el botón ``Flag''.

Aunque la interfaz gráfica facilita la interacción con el sistema conversacional, el usuario requiere abrir el notebook, ejecutar manualmente el código y las celdas de programación, considerando que se requiere la carga del LLM a través de un terminal interactivo, lo que demanda un mayor grado de intervención manual y manejo de comandos en Linux. Además, es posible que la sesión se interrumpa después de cierto tiempo o por inactividad. Por tanto, este entorno demanda conocimientos técnicos básicos y no ofrece un servicio continuo de ejecución. 

\subsubsection{Versión en Hugging Face}
Para que nuestro sistema conversacional sea realmente valioso y accesible, debe estar disponible públicamente y permanentemente a través de un servidor en Internet. La aplicación se desplegará como una página web interactiva, permitiendo a los usuarios ejecutarla desde cualquier navegador, sin requerir instalación alguna.

\textit{Hugging Face} \cite{ref36} es una plataforma de alojamiento y despliegue muy conocida en el campo de la IA, que ofrece planes de tipo gratuito y de pago. Tras registrarse en la plataforma, se procede a crear un \textit{Space} o repositorio para el proyecto. En este caso, el repositorio es público y dispone de una CPU en la modalidad gratuita. El nombre asignado al Space forma parte de la dirección web que puede compartirse para acceder y ejecutar la aplicación.
En el repositorio creado (denominado \textit{ragsql}) se incorpora el código junto con los archivos que siguen los lineamientos de despliegue de la plataforma. La estructura de la aplicación web se organiza de la siguiente manera:

\begin{verbatim}
ragsql/
|-- app.py             # archivo central de la aplicación
|-- requirements.txt   # lista de dependencias de Python
|-- tesis.db           # base de datos SQLite de tesis  
|-- README.md          # descripción del proyecto             
\end{verbatim}

Se trasladó el código del notebook al archivo principal \textit{app.py}, con adaptaciones necesarias para el nuevo entorno de ejecución; sin embargo, el núcleo del sistema (arquitectura, componentes y lógica de procesamiento) permanece inalterado.
Las modificaciones realizadas se sintetizan en dos aspectos: el LLM y la interfaz de usuario.
Tras realizar varias pruebas, fue necesario reemplazar Llama 3.1 con \textit{Gemini 2.5 Flash Lite} \cite{ref37}, debido a la disponibilidad y accesibilidad de los recursos computacionales en esta nueva plataforma.
A diferencia de Google Colab, donde manualmente se descarga, instala y ejecuta el LLM dentro del entorno con comandos en la terminal, en Hugging Face, el acceso al LLM se realiza mediante una clave API, la cual permite conectarse al modelo ya listo en la nube, sin necesidad de instalarlo y configurarlo localmente.

\begin{figure}[H]
\centering
\fbox{\includegraphics[width=0.95\linewidth]{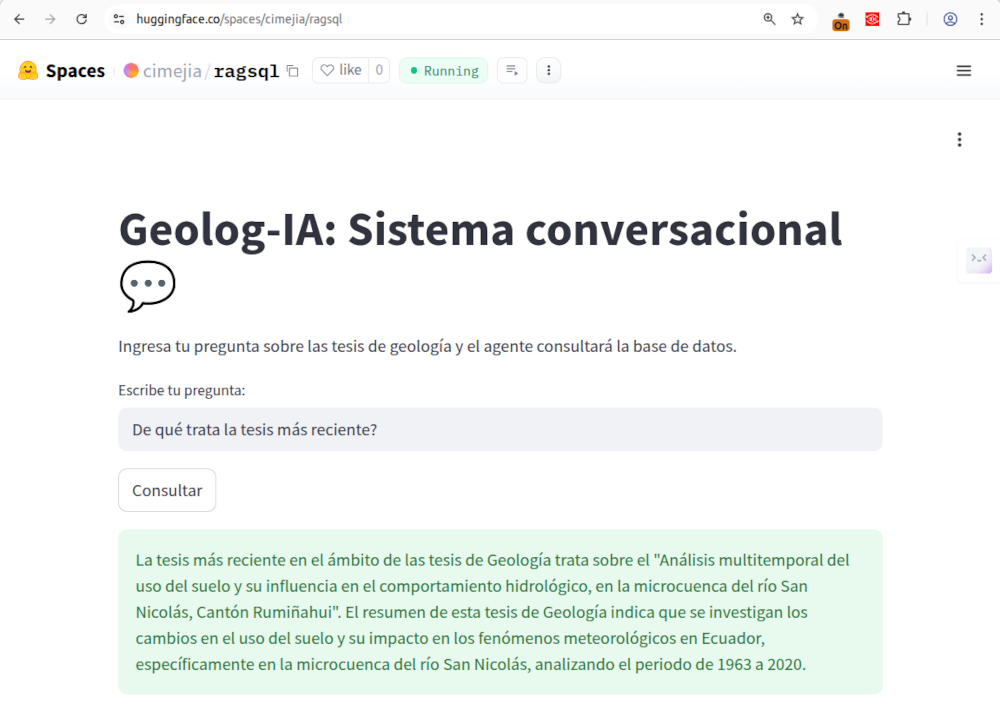}}
\caption{Interfaz gráfica del sistema conversacional creada en Hugging Face.}
\label{fig:webapp}
\end{figure}

Por otra parte, aunque Hugging Face permite el desarrollo de la interfaz gráfica de usuario con Gradio, se optó por \textit{Streamlit} \cite{ref8}. Esta librería de Python resultó ser la más adecuada según las necesidades específicas del sistema, facilitando la creación del encabezado, los títulos, los párrafos, los cuadros de texto y los botones de la página web, tal como se observa en la Figura \ref{fig:webapp}.

Cualquier usuario puede comenzar a utilizar la aplicación abriendo el enlace \url{https://huggingface.co/spaces/cimejia/ragsql} en su navegador web. Hugging Face se encarga de proporcionar la infraestructura computacional necesaria.

\section{Conclusiones}
\label{sec:conclusion}

El desarrollo del sistema conversacional Geolog-IA ha permitido superar las limitaciones de los sistemas tradicionales de búsqueda y recuperación de información.
Esto representa un avance significativo en la accesibilidad de las tesis académicas de geología para estudiantes, docentes y personal administrativo de nuestra institución.
La combinación de RAG, SQL y LLM ofrece a los usuarios una forma sencilla y eficiente de interactuar con la plataforma y extraer información detallada directamente en lenguaje natural.

Algunas acciones fueron determinantes para mejorar la organización y accesibilidad de la información.
A partir de la encuesta a los usuarios, se reestructuró la base de datos de tesis, descartando campos irrelevantes o redundantes, creando un nuevo campo (temática) y renombrando ``abstract'' a ``resumen''.
La selección del LLM consideró un equilibrio entre accesibilidad y buen desempeño.
Llama 3.1 y Gemini 2.5 demostraron notable capacidad de comprensión y generación de texto, equiparable a las opciones de pago.
Para asegurar la efectividad del agente, es fundamental complementar el prompt por defecto con un diseño estratégico del prompt de depuración para la corrección de errores, así como el prompt de respuesta para asegurar una salida final coherente y valiosa para el usuario.

Un reto importante es dar respuesta a preguntas tanto cuantitativas como cualitativas.
Geolog-IA soluciona ambos problemas con la base de datos estructurada, que es ideal para consultas cuantitativas, obteniendo estadísticas como \emph{cantidad de tesis aprobadas por año}  o \emph{número de tesis dirigidas por un tutor}.
En el caso de respuestas más interpretativas, el campo ``resumen'' permite la recuperación de información descriptiva y contextualizada.
De este modo, para una consulta como \emph{¿Qué tesis han abordado el tema de  riesgos volcánicos?}, el sistema no solo busca coincidencias en los títulos o palabras clave, sino que también analiza los resúmenes de las tesis para ofrecer respuestas más precisas y detalladas.
Esta capacidad le otorga un valor agregado dentro del ámbito académico.


El desempeño de Geolog-IA evaluado con la métrica BLEU (valor promedio de 0.87) indica que el sistema es capaz de proporcionar respuestas con alta coherencia y precisión.
Esto permitió implementar una aplicación web gratuita en dos modalidades: una dedicada al desarrollo y la experimentación, y otra orientada a la usabilidad práctica.
Esta última, con una interfaz accesible y de fácil uso, garantiza que una amplia comunidad académica pueda realizar consultas de manera intuitiva, sin importar su nivel técnico.
Por último, Geolog-IA no solo democratiza el conocimiento geológico en la universidad, sino que también establece un modelo de referencia para futuras aplicaciones en otros campos del conocimiento, sentando un precedente para el desarrollo de nuevas herramientas en otras disciplinas.


\end{document}